\documentclass[letterpaper]{article}

\usepackage{amsthm}
\usepackage{amsmath}
\usepackage{amssymb}
\usepackage{algorithmic}
\usepackage{hyperref}
\usepackage{fancybox}
\usepackage{graphicx}
\usepackage[in]{fullpage}

\hypersetup{
    colorlinks,
    citecolor=green,
    filecolor=black,
    linkcolor=blue,
    urlcolor=blue
}

\newcommand{\ve}{\mathbf{e}}

\newcommand{\vu}{\mathbf{u}}
\newcommand{\vv}[1]{\mathbf{#1}}

\newcommand{\vx}{\mathbf{x}}
\newcommand{\vy}{\mathbf{y}}

\newtheorem{property}{Property}

\newcommand{\argmax}[1]{\operatorname*{arg\,max}_{#1}}
\newcommand{\argmin}[1]{\operatorname*{arg\,min}_{#1}}

\newcommand{\ltriplebar}{\lvert\kern-0.9pt\lvert\kern-0.9pt\lvert}
\newcommand{\rtriplebar}{\rvert\kern-0.9pt\rvert\kern-0.9pt\rvert}

\newcommand{\bb}{\mathbf{b}}

\newcommand{\be}{\mathbf{e}}

\newcommand{\bx}{\mathbf{x}}
\newcommand{\by}{\mathbf{y}}

\newcommand{\bD}{\mathbf{D}}

\newcommand{\bI}{\mathbf{I}}

\newcommand{\bK}{\mathbf{K}}
\newcommand{\bL}{\mathbf{L}}
\newcommand{\bM}{\mathbf{M}}

\newcommand{\bR}{\mathbf{R}}

\newcommand{\bW}{\mathbf{W}}

\begin{document}

\title{Denoising: \Large{A Powerful Building-Block for Imaging, Inverse Problems,\\ and Machine Learning}}
\date{}

\author{Peyman Milanfar,  Mauricio Delbracio \\[0.5em]
Google\\
\small Mountain View, CA, USA \\
\small \texttt{\{milanfar,mdelbra\}@google.com}}

\maketitle

\begin{abstract}
Denoising, the process of reducing random fluctuations in a signal to emphasize essential patterns, has been a fundamental problem of interest since the dawn of modern scientific inquiry. Recent denoising techniques, particularly in imaging, have achieved remarkable success, nearing theoretical limits by some measures. Yet, despite tens of thousands of research papers, the wide-ranging applications of denoising beyond noise removal have not been fully recognized. This is partly due to the vast and diverse literature, making a clear overview challenging.

This paper aims to address this gap. We present a clarifying perspective on denoisers, their structure, and desired properties. We emphasize the increasing importance of denoising and showcase its evolution into an essential building block for complex tasks in imaging, inverse problems, and machine learning.  Despite its long history, the community continues to uncover unexpected and groundbreaking uses for denoising, further solidifying its place as a cornerstone of scientific and engineering practice.

\end{abstract}

\maketitle

\section{Introduction} \label{sec:intro}

Like most things of fundamental importance, image denoising is easy to describe, and very difficult to do well in practice. It is therefore not surprising that the field has been around since the beginning of the modern scientific and technological age - for as along as there have been sensors to record data, there has been noise to contend with.

Consider an image $\vx$, composed of a ``clean'' (smooth\footnote{It is important to note that this ``smooth'' component can contain edges and textures, hence we are using the term rather loosely here to describe operators that remove small-scale effects, leaving larger scale and higher contrast discontinuities alone.}) component $\vu$, and a ``rough'' or noisy component $\ve$, which we take to be zero-mean Gaussian white noise of variance $\sigma^2$, going forward: 
\begin{equation}\label{measurement_model}
\vv{x} = \vv{u} + \vv{e},
\end{equation}
where all images are scanned lexicographically into vectors.  The aim of any denoiser is to decompose the image $\vx$ back into its constituent components - specifically, to recover an estimate of $\vu$, the underlying signal, by applying some operator (denoiser) $f(\cdot,\alpha)$, parameterized by some $\alpha$ as follows: 
\begin{equation}
\vv{\Hat{x}(\alpha)} = f(\vv{x};\alpha) \approx \vu,
\end{equation}
where $\alpha(\sigma^2)$ is a monotonic function of the noise variance, and therefore controls the ``strength'' of the denoiser.

As the description above indicates, a denoiser is not a single operator but a {\em family} of bounded\footnote{We assume all images are in the numerical range [$0,1$]. In practice, an $8$-bit image would have values in [$0,255$] range.} maps $f(\vv{x},\alpha): \left[0,1\right]^N  \rightarrow \left[0,1\right]^N$. We expect ``good'' denoisers to have certain naturally desirable properties, which alas in practice, many do not. For the sake of completeness, and as a later guide for how to design good denoisers, we call a denoiser {\em ideal} if it satisfies the following properties:

\begin{property}{(Identity)} \label{prop:identity} When there is no noise (i.e. $\alpha = 0$), the ideal denoiser will reproduce the input unchanged.\\
\begin{equation}
    f(\vx,0) = \vx, \;\;\;\;\;\; \forall \vx. 
\end{equation}
That is, $f(\vv{x},0)$ is the identity operator.
\end{property}

\begin{property}{(Conservation)}\label{prop:invariance}
An ideal denoiser has a symmetric Jacobian\footnote{Unless explicitly noted otherwise, $\nabla$ will mean $\nabla_{\vx}$ throughout the paper.}
\begin{equation}
    \nabla f(\vx,\alpha) = \nabla f(\vx,\alpha)^T.
\end{equation}
Or equivalently, 
\begin{equation}
    f(\vx,\alpha) = \nabla \mathcal{E}(\vx,\alpha),
\end{equation} 
for some scalar-valued, differentiable (``potential'' or ``energy'') function $\mathcal{E}(\vx,\alpha)$. This also means that the ideal denoiser defines a conservative vector field\footnote{We note that since the ideal denoiser can be expressed as the gradient of a scalar function, this leads directly to the path independence property of line integrals, which is the defining characteristic of a conservative vector fields.}. 
\end{property}

To convey some intuition for this property, consider the linear case. When a denoiser is linear: $f(\vx,\alpha) = W(\alpha)\vx$, we always require the matrix $W(\alpha)$ to be row-stochastic (meaning the rows sum to $1$) in order to preserve the mean local brightness. Ideally, we also require $W(\alpha)$ to be symmetric~\cite{milanfar2013symmetrizing}, which has the added advantage that the denoiser is \textit{admissible}~\cite{cohen1968} in the mean-square sense. Property $2$ extends these notions to more general nonlinear denoisers \footnote{It's worth noting that the combination of symmetric and row-stochastic implies that $W(\alpha)$ is doubly-stochastic.}.

\paragraph{Remark:} The conservation Property 2 guarantees that the ideal denoiser is the gradient of a scalar field.This also implies that $f(\vx,\alpha)$ is a Lipschitz map with some constant $M(\alpha)$: 
\begin{equation}
    \|f(\vx,\alpha) - f(\vy,\alpha)\| \leq M(\alpha) \|\vx - \vy \|.
\end{equation} 

We naturally expect  $f(\mathbf{0},\alpha) = \mathbf{0}$ for all $\alpha$; therefore, this Lipschitz condition implies $\|f(\vx,\alpha)\| \leq M(\alpha) \|\vx \|$. A non-expansive denoiser would require that $M(\alpha)\leq 1$. In the statistics literature, such operators are called \textit{shrinking} smoothers~\cite{hastie2009elements,buja1989linear}. 
 
\vspace{0.2in}
The above properties impose the structure of an \textit{affine} space~\cite{berger} on the class of ideal denoisers. Namely, any affine combination of ideal denoisers is also ideal. That is, if we let\footnote{Note that we do not place a constraint on the sign of $a_k$'s.}
\begin{equation}
    g_a(\vx,\alpha) = \sum_{k=0}^N a_k \: f(\vx,\alpha_k) \;\;\;\;\; \text{with}
    \;\;\;\;\; \sum_{k=0}^N a_k = 1,
\end{equation}
\noindent it is easy to verify that Properties $1$ and $2$ are satisfied. 
\vspace{1em}
\hrule
\vspace{1em}
\textbf{Summary:} Ideal denoisers satisfy: 
\begin{itemize}
\item \textbf{Property 1:} \;\; $f(\vx,0) = \vx$, 
\item \textbf{Property 2:} \;\; $f(\vx,\alpha) = \nabla \mathcal{E}(\vx,\alpha)$, 
\item Closed-ness under affine linear combination.
\end{itemize}
\hrule
\vspace{1em}

It is an unfortunate fact that in practice, most denoisers are not ideal. But this should not bother the reader, as by studying the broader class of denoisers we will learn how the above desirable properties are manifested or desired in practice, and which practical denoisers (approximately or exactly) satisfy them.\vspace{0.5em}

\noindent \textbf{A note on this work:} Rather than a survey of image denoising, this work focuses on defining ideal denoisers, their properties, and their connections to statistical theory and machine learning. We then demonstrate how these powerful components can serve as building blocks in various applications.  Readers interested in a historical overview of image denoising are encouraged to consult the excellent resources in  ~\cite{lebrun2012secrets,milanfar2013tour,bertalmio2018denoising,tian2020deep,elad2023image}.
Our analysis specifically considers an additive white Gaussian noise model due to its broad applicability and relevance to the applications explored herein.  A deeper examination of various noise models can be found in~\cite{hasinoff2010noise,lebrun2012secrets,aguerrebere2013study}.

\section{Denoising as a Natural Decomposition}
One of the remarkable aspects of well-behaved (even if not ideal) denoising operators is that we can employ them to easily produce a natural multiscale decomposition of an image, with perfect reconstruction property\footnote{This is similar in spirit to the classic multiscale decomposition in~\cite{burt1987laplacian}, except that there is no decimation, and the filters are nonlinear here.}. To start, consider a denoiser $f(\vx,\alpha)$. We can write the obvious relation: 
\begin{equation}
\vx = f(\vx,\alpha) + \left[ \vx - f(\vx,\alpha) \right].
\label{eq:natural_decomposition}
\end{equation}
The first term on the right-hand side is a {\em smoothed} (or denoised) version of $\vx$, whereas the second term in the brackets is the residual $r_0(\vx,\alpha) = \vx - f(\vx,\alpha)$ which is an ostensibly ``high-pass'' version. Next, we can apply the same decomposition repeatedly to the already-denoised components\footnote{To simplify the exposition, we use the same denoiser and the $\alpha$ at each step, but this is not necessary.}: 
\begin{eqnarray}
\vx & = & f(f(\vx,\alpha),\alpha) + \left[f(\vx,\alpha)- f(f(\vx,\alpha),\alpha) \right] + r_0(\vx,\alpha) \nonumber\\
& = & f(f(\vx,\alpha),\alpha) + r_1(\vx,\alpha) + r_0(\vx,\alpha) \nonumber\\
& \vdots & \nonumber\\
& = & f^n(\vx,\alpha) + \sum_{k = 0}^{n-1} r_k(\vx,\alpha),
\end{eqnarray}
\noindent where $f^n$ denotes the operator applied $n$ times (i.e. a diffusion process), and $r_k = f^k - f^{k+1}$ (i.e. a residual process). For any $n$, this $n$-th order decomposition splits $\vx$ {\em exactly} into a smooth component $f^n(\vx,\alpha)$ and a sequence of increasingly fine-detail components $r_k(\vx,\alpha)$. 

It is important to note that applying the operators $f(\vx,\alpha)$ multiple times does not necessary result in a completely smooth result. For instance, if we repeatedly apply a bilateral filter~\cite{tomasi1998bilateral,milanfar2013tour}, the result is a {\em piece-wise} constant image. The process we've described here has been called, in certain instances, a \textit{cartoon-plus-texture} decomposition in
~\cite{cartoon,Osher05aniterative}, mainly in the context of total-variation denoising. Our point of view is considerably more general, applicable to {\em any} denoiser. 

Returning to the decomposition above, it empowers us to do practically useful things. For instance, truncating the residual terms at some $n$, we can smooth out certain high frequency features. More generally, we can null out any component in the sum; or better yet, recombine the components with new coefficients to produce a \textit{processed} or modified image, as follows: 
\begin{equation}
    g_a(\vx,\alpha,\beta) = \beta_n f^n(\vx,\alpha) + \sum_{k = 0}^{n-1} \beta_k r_k(\vx,\alpha). 
\end{equation}

\begin{figure}[t]
    \centering
    \includegraphics[width=\linewidth]{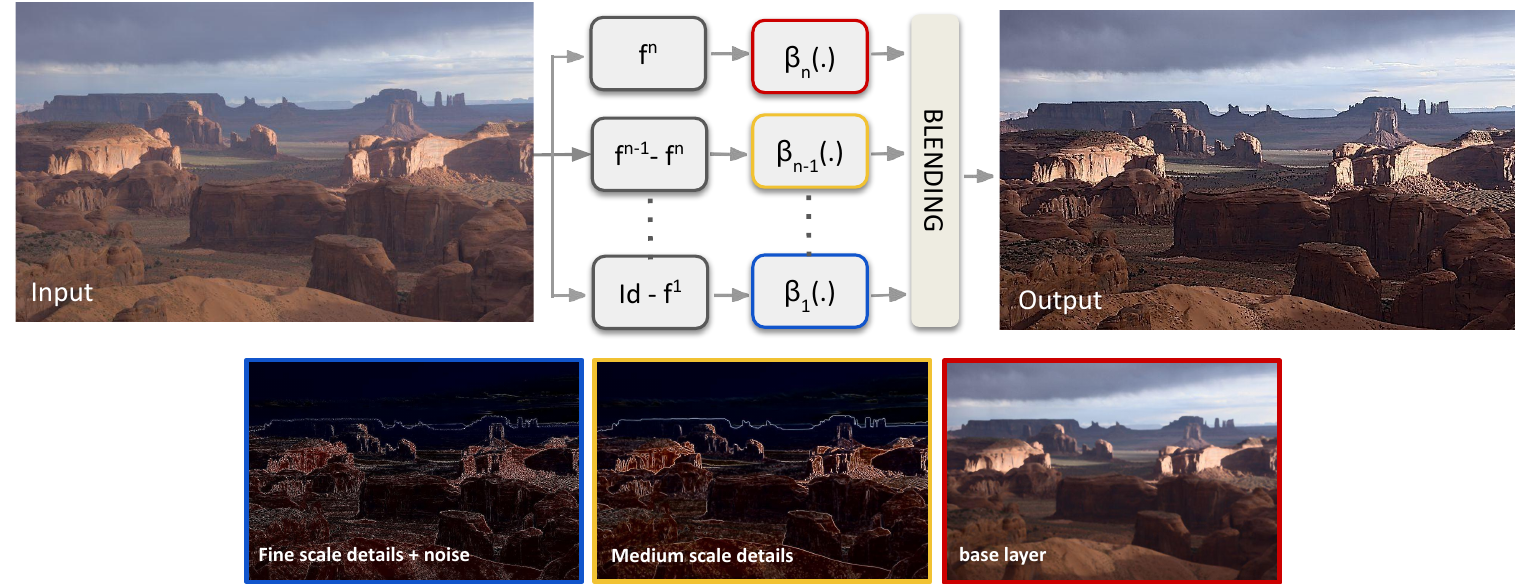}
    \caption{Denoising as a natural image decomposition. Image adapted from~\cite{talebi2016fast}.}
    \label{fig:denoising-multiscale-decomposition}
\end{figure}

This approach was generalized and used in a practical setting in~\cite{talebi2016fast,talebi2014nonlocal} to produce a wide variety of image processing effects, built on a base of well-established (at the time) non-local means denoisers. This is illustrated in Figure~\ref{fig:denoising-multiscale-decomposition}. 
More generally, given paired examples of input and desired output images ($\vx_i,\widehat{\vx}_i$), one can construct a loss function such as shown below, where $d$ is a training loss, and $\mathcal{R}$ is a regularization term. By minimizing this loss, we can learn both the parameters $\alpha$ and $\beta$. 
\begin{equation}
    \text{Loss}_f(\alpha, \beta) = \frac{1}{N} \sum_{i=1}^{N} d(\widehat{\mathbf{x}}_i, g_\beta(\mathbf{x}_i, \alpha)) + \mathcal{R}(\alpha, \beta).
\end{equation}
Recently, in~\cite{geng2024factorized} the authors used a similar decomposition to create a zero-shot method to control each individual component of the decomposition through diffusion model sampling.

\paragraph{Connection to Residual Networks:}
The concept of breaking down an image into layers of varying detail is closely related to the architecture of Residual Neural Networks~\cite{he2016deep} (ResNets). Both share the principle that it's simpler to model/learn residual mappings (the difference between the input and desired output) than to model/learn the complete transformation directly. While traditional deep neural networks try to learn this complex mapping in one go, ResNets use ``skip connections'' that allow the network to bypass layers, adding the original input to a later layer's output. Letting $H(\vx)$ be the desired complex mapping and $\vx$ the input, a ResNet layer attempts to learn a residual function $F(\vx)$ such that:
\begin{equation}
H(\vx) = \vx + F(\vx),
\end{equation}
The skip connection ensures that the original input $\vx$ is preserved and added back to the output. Note the connection to~\eqref{eq:natural_decomposition}, where the residual term is exactly $-F(\vx)$.
This decomposition and the use of skip connections simplify the network's task, making optimization easier and mitigating the vanishing gradient problem that can hinder deep network training~\cite{weinan2017proposal}. Additionally, the preservation of the original input or its smooth approximation through skip connections ensures important information isn't lost as data travels through the network\footnote{There are actually ways of ensuring invertibility of ResNets, see e.g.,
~\cite{pmlr-v97-behrmann19a}.}.  ResNets have been a major breakthrough in deep learning, enabling the training of much deeper networks and achieving state-of-the-art performance on image recognition tasks, with the concept of residual learning now being applied to other domains beyond image processing.

\paragraph{Image denoisers for anomaly detection:} The natural decomposition of an image using denoisers has also been used for analyzing images, for example to detect anomalies~\cite{davy2018reducing,ehret2019image} in images. The principle behind this is that anomalies, being infrequent occurrences, lack the self-similarity or smoothness typically observed in natural images. Drawing inspiration from patch-based denoising (e.g., non-local means), which employs self-similarity to differentiate between signal and noise, in~\cite{davy2018reducing} the authors introduce a method that effectively dissects an image into two components. The first is a self-similar component that embodies the background or 'smooth' regions of the image given by the denoiser. The second is a residual component that encapsulates the distinctive, non-repetitive elements, which could potentially include anomalies and noise. The residual image, anticipated to resemble noise, is then subjected to a statistical test to detect any anomalies.

Next, we will describe various well-known classes of denoisers, including those derived from statistical optimality principles, and others which are pseudo-linear and derived from non-parametric or empirical considerations. We will also examine whether these classes of denoisers satisfy the above properties. 

\section{The Structure of General Denoisers}

\subsection{Bayesian Denoisers}
Bayesian denoising invokes the use of a prior $P(\vu)$ on the class of ``clean''
images $\vu$. This prior influences the estimate of the underlying signal away
from the observed measurement $\vx$. We will describe the popular Maximum a-Posteriori (MAP) and the Minimum Mean-Squared Error (MMSE) denoisers below. 

\begin{figure}[th]
    \centering
    \includegraphics[width=0.55\linewidth]{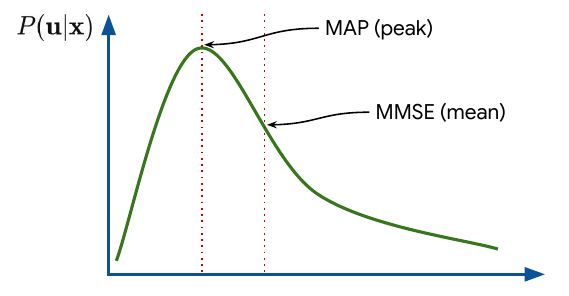}
    \caption{Bayesian Denoisers: MAP vs.\ MMSE.}
    \label{fig:BayesEstimators}
\end{figure}

The contrast between the MAP and MMSE is highlighted in Figure~\ref{fig:BayesEstimators}. The two estimates tend to coincide when the posterior is
symmetric and unimodal, or when the noise variance $\sigma^2$ is small. 

\subsubsection*{Maximum a-Posteriori (MAP)}
As the name indicates, the maximum a posteriori estimate is the value of $\vu$ at which the posterior density $P(\vu|\vx)$ is maximized,
\begin{equation}
\vv{\Hat{x}}_\mathit{map} = \argmax{\vu}P(\vu|\vx).
\end{equation}
When the noise is Gaussian and white, the optimization
boils down to regularized least-squares
\begin{equation}\label{e:mapminimization}
\vv{\Hat{x}}_\mathit{map}
= \argmin{\vu}\frac{1}{2}\|\vu-\vx\|^2  + \alpha \: \phi(\vu),
\end{equation}
where $\phi(\vv{u}) = -\log P(\vv{u})$ is the negative log-prior on the space of ``clean'' images, and $\alpha$ is proportional to the noise variance. It would appear that the MAP denoiser does not have a closed form. However, the expression \eqref{e:mapminimization} is also known in the optimization literature as a \emph{proximal operator}~\cite{moreau1965proximite,gabay1976dual} when $\phi$ is convex, quasi-convex, or a difference of convex functions. It is well-known
~\cite{moreau1965proximite,gribonval2020} that to every proximal operator $f$ there corresponds a convex scalar-valued function $\psi$ such that $f = \nabla \psi$.

Furthermore, in the context of the MAP estimate, $\psi$ has an explicit form:
\begin{equation} \label{eq:MAP_Shrinkage}
\begin{aligned}
\psi(\vx) & = \left[\frac{1}{2}\|\vx\|^2 - \alpha \phi_\alpha(\vx)\right] 
\;\; \implies \;\; \Hat{\vx}_\mathit{map} = \nabla \psi(\vx) = \vx - \alpha \nabla \phi_\alpha(\vx),
\end{aligned}
\end{equation}
where $\phi_\alpha$ is a {\em smoothed} version of $\phi$ called its Moreau
envelope~\cite{moreau1965proximite,boyd2011distributed,proximal2015,burger2016first}. As we will see below, the MMSE estimate shares a very similar form. 

\begin{figure}
    \centering
    \includegraphics[width=0.75\linewidth]{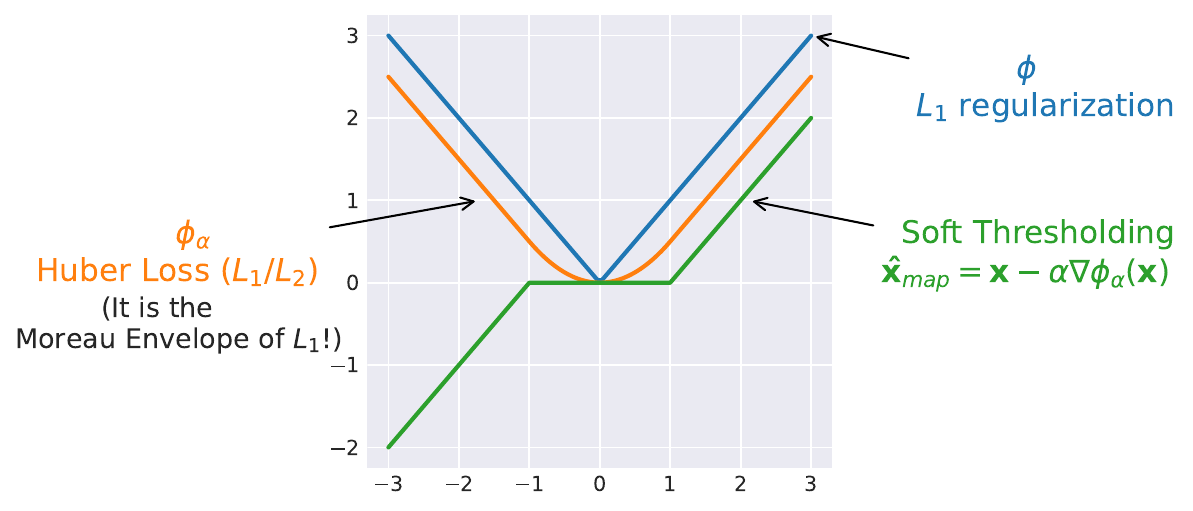}
    \caption{Example of MAP denoiser with $L_1$ loss, with $\alpha = 1$. The Moreau envelope is the Huber loss.}
    \label{fig:L1-Huber}
\end{figure}

An example (for the scalar case) of the MAP denoiser for $\phi(\cdot)=\|\cdot\|_1$ is shown in Figure~\ref{fig:L1-Huber}, where the resulting denoiser is exactly the soft-thresholding operator. 

\subsubsection*{Minimum Mean-Squared Error Denoising}
While Maximum A Posteriori (MAP) denoising seeks the most probable estimate of a clean signal given a noisy observation, MMSE denoising aims to find the estimate that minimizes the mean squared error (MSE) between the estimate and the true signal
\begin{equation}
MSE(\hat{\vx},\vu) =\mathbb{E}_{\vu, \vx } \left[ \| \hat{\vx} - \vu \|^2_2 \right],
\end{equation}
where $\vu$ is the true signal, $\vx$ the noisy observation, and $\Hat{\vx}$ is the estimate of $\vu$ given $\vx$.

\vspace{.5em}
\noindent \textbf{The Posterior Mean as the MMSE Estimator.}
A fundamental result in estimation theory is that the posterior mean, $\mathbb{E}\left[\vu|\vx\right]$, is the MMSE estimator. This can be shown by minimizing the MSE directly. Starting with the definition of MSE:
\begin{align}
\text{MSE} &= \mathbb{E}_{\vu,\vx} \left[ \| \hat{\vx} - \vu \|^2_2 \right]  \label{eq:mse}\\
&= \int \left[ \int \| \hat{\vx} - \vu \|^2_2 \, P(\vu | \vx) \, d\vu \right] P(\vx) \, d\vx.
\end{align}
Since $P(\vx) \ge 0$, minimizing the MSE is equivalent to minimizing the inner integral for each $\vx$. Expanding the square and simplifying, we get:
\begin{align}
\int \| \hat{\vx} - \vu \|^2_2 \, P(\vu | \vx) \, d\vu &=  \hat{\vx}^T\hat{\vx} - 2 \hat{\vx}^T \int \vu \, P(\vu | \vx) \, d\vu + \int \vu^T\vu \, P(\vu | \vx) \, d\vu.
\end{align}
Taking the derivative with respect to $\hat{\vx}$ and setting it to zero, we find:
\begin{align}
\hat{\vx} &= \int \vu \, P(\vu | \vx) \, d\vu = \mathbb{E}\left[\vu|\vx\right].
\end{align}
Thus, the posterior mean minimizes the MSE for any $\vx$, and therefore minimizes the overall MSE.

\begin{figure}[htpb]
\begin{minipage}{.33\textwidth}
    \centering
    \includegraphics[width=.85\linewidth]{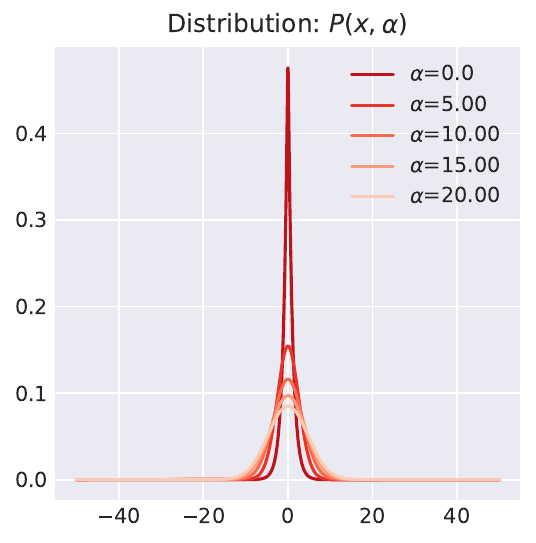}\\
    
\end{minipage}
\begin{minipage}{.33\textwidth}
\centering
\includegraphics[width=.85\linewidth]{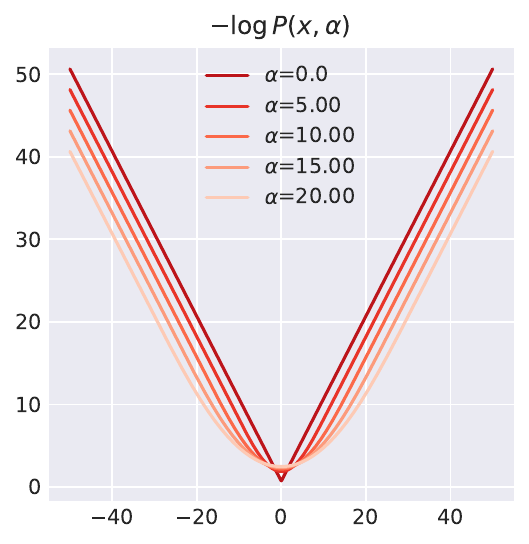}

\end{minipage}
\begin{minipage}{.33\textwidth}
\centering
\includegraphics[width=.85\linewidth]{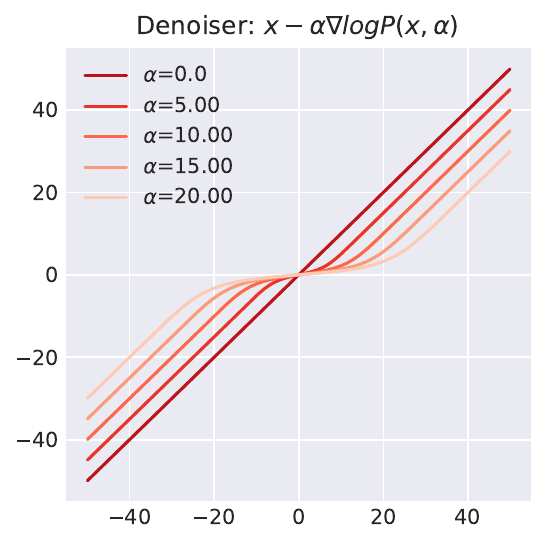}

\end{minipage}
\caption{
Illustration of a one-dimensional MMSE denoiser employing $L_1$ regularization, demonstrating the impact of varying $\alpha$. The visualization progresses from the smoothed distribution $P(\vx, \alpha)$ (left), to the corresponding Energy function (middle), and ultimately, the resulting denoiser (right).}
\label{fig:L1distribution}
\end{figure}

\begin{figure}[ht]
\begin{minipage}{.585\textwidth}
    \centering
    \includegraphics[width=\linewidth]{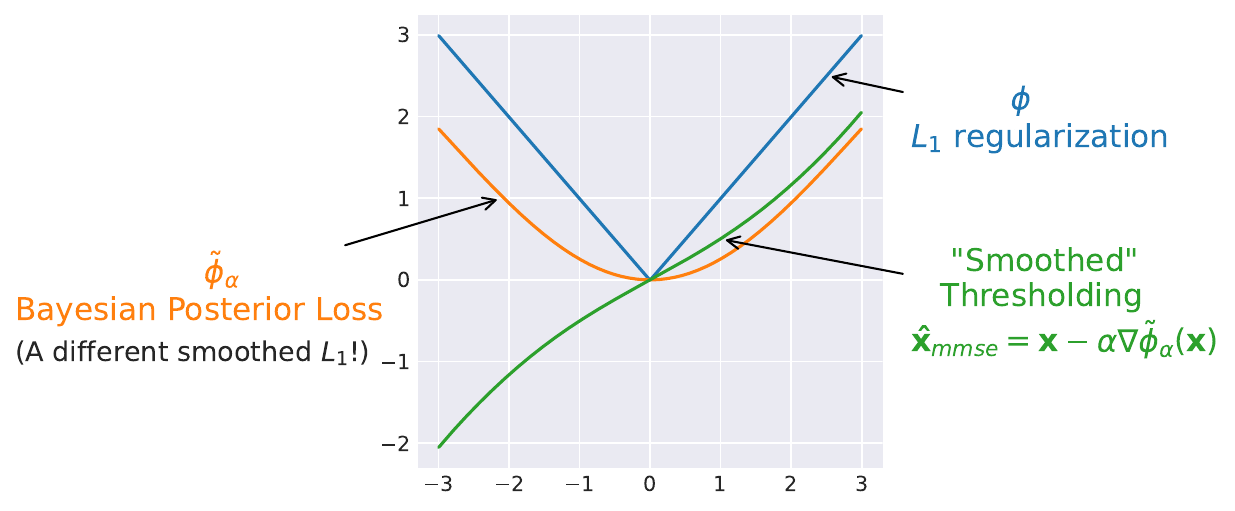}\\
    
\end{minipage}
\begin{minipage}{.405\textwidth}
\centering
\includegraphics[width=\linewidth]{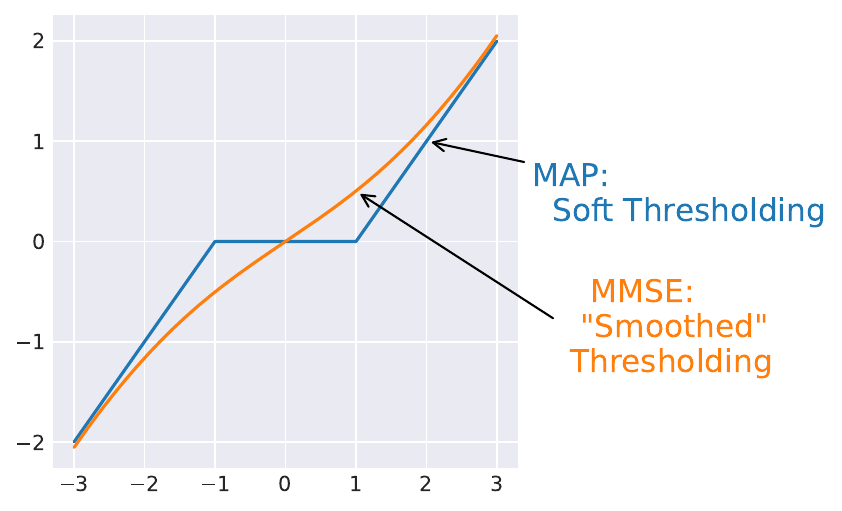}

\end{minipage}
\caption{Left: Example of MMSE denoiser with $L_1$ loss, with $\alpha = 1$. Right: Comparison of MAP and MMSE denoiser for the $L1$ loss, with $\alpha=1$.}
\label{fig:SmoothedThreshold}
\end{figure}

\vspace{.5em}
\noindent \textbf{Tweedie's Formula and the MMSE Denoiser:}
While the MMSE expectation integral is generally difficult or impossible to evaluate directly, a key result known as Tweedie's formula~\cite{robbins56,stein81,efron11} enables us to write the expression for MMSE also in the form of the gradient of a scalar function: 
\begin{equation}\label{eq:tweedie}
\begin{aligned}
\vv{\Hat{x}}_\mathit{mmse} & = \mathbb{E}(\mathbf{u}|\mathbf{x})
= \vx + \alpha \nabla \log P(\vx,\alpha) = \nabla\left[\frac{1}{2}\|\vx\|^2 + \alpha \log P(\vx,\alpha)\right],
\end{aligned}
\end{equation}
where $\alpha = \sigma^2$ and $P(\vx,\alpha)$ is the marginal density of the measurement $\vx$, computed as $P(\vv{x},\alpha) = \int P(\vv{x}|\vv{u},\alpha) P(\vv{u}) \,d\vv{u}$. 
It is apparent that $P(\vx,\alpha)$ is effectively the prior $P(\vv{u})$ {\em blurred} with the noise distribution (Gaussian in our setting). 
Just like the MAP denoiser, the MMSE denoiser also has the form $f(\vx) = \nabla \tilde\psi$. More specifically, the MMSE denoiser can be rewritten as
\begin{equation} \label{eq:MMSE_Shrinkage}
\vv{\Hat{x}}_\mathit{mmse} = \vx - \alpha \nabla \tilde{\phi}_\alpha(\vv{x}),
\end{equation}
where $\tilde{\phi}_\alpha(\vv{x}) = -\log P(\vv{x},\alpha)$. This is more or less identical to the form of the MAP denoiser in~\eqref{eq:MAP_Shrinkage}.
Figure~\ref{fig:L1distribution} illustrates the MMSE denoiser for the scalar case with $L_1$ penalization, showcasing its behavior across various $\alpha$ values.  A comparison between the MMSE and Maximum A Posteriori (MAP) estimators is presented in Figure~\ref{fig:SmoothedThreshold}.

\vspace{0.2in}

\noindent \textbf{Data-driven MMSE denoisers:}
The typical modern supervised approach to image denoising is to train a deep neural network with pairs of clean and noisy images, where the noise is often modeled as additive white and Gaussian(AWGN)~\cite{elad2023image}. 
Let's assume we have image pairs $(\vu,\vx) \sim P(\vu, \vx)$ where $\vu$ represents a clean image, and $\vx$ is the noisy observation obtained by adding AWGN with a known standard deviation to $\vu$: $\vx = \vu + \ve$, where $\ve\sim \mathcal{N}(0,\sigma^2\bI)$.

A typical regression approach would predict $\vu$ directly from $\vx$ using a trained model $\hat{\vx} = F_\theta(\vx) \approx \vu$, by minimizing the expected reconstruction error:
\begin{equation}
\min_\theta \mathbb{E}_{\vu,\ve} \| F_\theta(\vu + \ve) - \vu \|^p_p \approx \min_\theta \sum_i \| F_\theta(\vu^i + \ve^i) - \vu^i \|^p_p.
\end{equation}
In the case $p=2$, this leads to an approximation to the ideal MMSE denosier,
$ \vx_{\text{MMSE}} = \mathbb{E}\left[\vu|\vx\right] = \int \vu\, P(\vu \mid \vx) d\vu.$

As mentioned earlier, the MMSE denoiser is the average of all plausible clean signals given the noisy observation. This averaging can lead to a loss of details and a blurry appearance, especially when the noise level is high. 
This is because minimizing average distortion (e.g., PSNR) can harm perceptual quality~\cite{blau2018perception}. To address this, alternatives including perceptual~\cite{johnson2016perceptual,zhang2018unreasonable,delbracio2021projected} and adversarial losses~\cite{goodfellow2014generative,ledig2017photo,isola2017image} have been considered. A more powerful approach is to \emph{sample} from the posterior distribution, avoiding the \emph{regression to the mean} effect
~\cite{kawar2021stochastic,kawar2021snips,kawar2022denoising,ohayon2021high,kadkhodaie2021stochastic,li2021srdiff,saharia2022image,delbracio2023inversion,ohayon2024posterior}.

Denoising Autoencoders (DAEs) are a prime example of data-driven MMSE denoisers~\cite{vincent2008extracting}. These neural networks excel at learning robust data representations by training on noisy input and striving to reconstruct the original, clean data. This makes them not only valuable for denoising but also for tasks like data compression and feature extraction.

\subsection{Energy-based Denoisers}\label{sec:energy-denoisers}

We've seen that both MMSE and MAP estimators are of the form $f(\vx,\alpha) = \vx - \alpha \nabla \phi_\alpha(\vx)$ where $\phi_\alpha(\vx)$ is some smoothed version of $\phi$ -- differently smoothed in each case. These denoisers are also special cases of a general ``energy-based'' formulation~\cite{cohen2021}:
\begin{equation}
f(\vx,\alpha) = \nabla \mathcal{E}(\vx,\alpha),
\end{equation}
where in the particular case of MMSE and MAP,
\begin{equation}
\mathcal{E}(\vx,\alpha) = \frac{1}{2}\|\vx\|^2 - \alpha \phi_\alpha(\vv{x}).
\end{equation}

If the energy function satisfies $\nabla \mathcal{E}(\vx,0) = \vx$ for all $\vx$ (as do both the MMSE and MAP), then such denoisers are ideal. This is because the Jacobian of the denoiser can be written as 
\begin{equation}
\nabla f(\vx,\alpha) = \mathcal{H} \left[ \mathcal{E}(\vx,\alpha) \right],
\end{equation}
where $\mathcal{H}$ denotes the Hessian operator which is, by definition, symmetric. In summary, all energy-based denoisers, including MAP and MMSE, are ideal, have symmetric Jacobians and are therefore conservative vector fields. 

\subsubsection*{Approximation of Energy-based Denoisers} 
The energy-based formulation of denoisers provides a natural mechanism for principled empirical design of denoisers. This approach turns out to be consistent with the well-established \emph{empirical Bayes}~\cite{robbins56,efron2012large} approach as well. 

Recall that the formulation of the denoising problem in \eqref{measurement_model} directly implies that the marginal density $P(\vx,\alpha)$ is smooth because  $P(\vx,\alpha) =  P \circledast \mathcal{N}(0,\alpha) (\vx)$, where $\circledast$ denotes convolution (i.e. blurring) with a Gaussian density. So by definition, this marginal density can be treated as a smooth function - the larger the noise parameter $\alpha$, the smoother is $P(\vx,\alpha)$.

Now let's write this marginal density in \emph{Gibbs} form:

\begin{equation}
P(\vx,\alpha)=\frac{1}{Z}\exp{\left[-\mathcal{E}(\vx,\alpha)\right]}.
\end{equation}

\noindent where $\mathcal{E}(\vx,\alpha)$ is an energy function with $\mathcal{E}(\mathbf{0},\alpha) =0$. The \emph{score function} is related to the energy function as follows: 
\begin{equation}
    s(\vx,\alpha) = \nabla \log P(\vx,\alpha) = - \nabla \mathcal{E}(\vx,\alpha).
\end{equation} 

The smoothness of $P(\vx,\alpha)$ implies smoothness of the energy $\mathcal{E}(\vx,\alpha)$, thereby ensuring the existence of the gradient for both. 

We can expand the energy around $\vx=0$ using a first order Taylor expansion (with the Lagrange form of the remainder) to get
\begin{equation}
    \begin{aligned}
        \mathcal{E}(\vx,\alpha) &= \mathcal{E}(0,\alpha) + \nabla \mathcal{E}(0,\alpha) \vx + \frac{1}{2}\vx^T \mathbf{L}(\mathbf{a},\alpha) \vx \\
 & = \nabla \mathcal{E}(0,\alpha) \vx + \frac{1}{2}\vx^T \mathbf{L}(\mathbf{a},\alpha) \vx,
    \end{aligned}
\end{equation}
where $\bL(\textbf{a},\alpha)$ represents the (symmetric) Hessian matrix of $\mathcal{E}$ evaluated at some (unknown) point $\mathbf{a}$ lying on the line segment\footnote{As such, the point $\mathbf{a}$ depends indirectly on $\vx$.} between $0$ and $\vx$.
Accordingly, the score function is
\begin{equation}
    s(\vx,\alpha)  = - \nabla \mathcal{E}(\vx,\alpha) = -\nabla \mathcal{E}(0,\alpha) - \mathbf{L}(\mathbf{a},\alpha) \vx.
\end{equation} 
Meanwhile, Tweedie's formula implies that the MMSE denoiser has the form:
\begin{equation}
    \begin{aligned}
f(\vx,\alpha) & = \vx + \alpha \: s(\vx,\alpha) \\
& = \vx - \alpha \nabla \mathcal{E}(0,\alpha) - \alpha \mathbf{L}(\mathbf{a},\alpha) \vx.
    \end{aligned}
\end{equation}
Requiring that $f(0,\alpha) = 0$ implies that the second term must be zero. Therefore, the MMSE denoiser has a simple (pseudo\footnote{Denoisers of this form are \emph{pseudo-linear}~\cite{milanfar_SPM2011} as they are similar in form to linear filters, except that the matrix $\bL$ implicitly depends on $\vx$.})-linear form: 
\begin{equation}
 \vv{\Hat{x}}_\mathit{mmse} = \vx - \alpha \mathbf{L}(\mathbf{a}) \vx = \left[\bI - \alpha \mathbf{L}(\mathbf{a},\alpha)\right] \vx.
\end{equation}

To summarize, the resulting locally optimal denoiser can be written as 
\begin{equation} \label{eq:pseudo-linear}
    f(\vx,\alpha) = \mathbf{W}(\vx,\alpha) \: \vx,
\end{equation}
where the symmetric matrix $\mathbf{W}(\vx,\alpha)$ is adapted to the structure of the input $\vx$. This observation is consistent with earlier findings~\cite{milanfar2013tour,kadkhodaie2024,srinivas2019full} that such pseudo-linear filters -including those built from (bias-free) deep neural nets- are (a) attempts at empirical approximations of the optimal MMSE denoiser, (b) shrinkage operations in an orthonormal basis adapted to the underlying structure of the image, and (c) perturbations of identity. In particular, such denoisers can be written in the form $f(\vx) = \nabla f(\vx) \vx$, meaning that their local behavior is fully determined by their Jacobian, and therefore its spectrum. 

Though these facts were neither historically clarified, nor the original motivation for their development~\cite{milanfar_SPM2011}, denoisers of the form~\eqref{eq:pseudo-linear} have always been heuristic/empirical approximations to the MMSE. These denoisers were hugely popular and effective (e.g.~\cite{bm3d,tomasi1998bilateral}) for decades before the more recent introduction of neural networks. More recent work by Scarvelis et al.
~\cite{scarvelis2023closed} explores the use of a specific kernel approach to create a ``closed-form'' diffusion model that operates directly on the training set, without the need for training a neural network.

Next, we will describe these types of denoisers -using the language of kernels- in more detail.

\subsection{Kernel Denoisers} \label{sec:kernel}

\paragraph{Motivation:} The basic idea behind kernel denoisers follows a non-parametric approach to modeling the distribution of (clean) images. Concretely, consider our basic setting given by
$$
\vv{x} = \vv{u} + \vv{e},
$$
where $\ve$ is zero-mean Gaussian white noise of variance $\alpha = \sigma^2$. In practice the density $P(\vu)$ is unknown, but we may have access to examples\footnote{Without loss of generality, $\vu_i$ can refer to either full images, or patches thereof.} $\vu_i$, for $i=1,\cdots,n$. We can construct a naive empirical estimate of the distribution as follows:
\begin{equation}
\widehat{P}(\vu) = \frac{1}{n} \sum_{i=1}^{n} \delta(\vu - \vu_i).
\end{equation} 
The empirical density for $\vx$ is the convolution of $\widehat{P}(\vu)$ with the Gaussian density $\mathcal{N}(0,\alpha \bI)$, yielding: 
\begin{equation}
\widehat{P}(\vx) = \frac{1}{n} \sum_{i=1}^{n} \mathcal{N}(\vx - \vu_i,\alpha \bI).
\end{equation}
Armed with this estimate, we can compute an empirical estimate of the score: 
\begin{equation}
    \nabla \log \widehat{P}(\vx) = -\frac{1}{\sigma^2} \left[\vx - \frac{\sum_{i}\vu_i \mathcal{N}(\vx - \vu_i,\alpha \bI)}{\sum_{i} \mathcal{N}(\vx - \vu_i,\alpha \bI)} \right].
\end{equation}
Invoking Tweedie's formula, we have a closed form approximation to the MMSE denoiser as a (data-dependent) weighted average~\cite{hastie2009elements} of the clean data points $\vu_i$:
\begin{equation}
\begin{aligned}
\widehat{\vx}_{mmse} & \approx \vx + \sigma^2 \nabla \log \widehat{P}(\vx) \\
& = \vx - \left[\vx - \frac{\sum_{i}\vu_i \mathcal{N}(\vx - \vu_i,\alpha \bI)}{\sum_{i} \mathcal{N}(\vx - \vu_i,\alpha \bI)} \right] \\
&  = \frac{\sum_{i}\vu_i \mathcal{N}(\vx - \vu_i,\alpha \bI)}{\sum_{i} \mathcal{N}(\vx - \vu_i,\alpha \bI)} \\
&  = \sum_{i}\vu_i W(\vx - \vu_i,\alpha \bI).
\end{aligned}
\end{equation}
In practice, we may only have access to the noisy image $\vx$.  In this scenario, we can treat each pixel $x_i$ as an independent sample (with independent noise) and apply the same reasoning directly to the noisy input, using it as a proxy for the clean signals:
\begin{equation}
\hat{x}_i  =  \sum_{i} x_j W(x_j - x_i, \alpha).
\end{equation}
This is a primitive instance of the pseudo-linear form alluded to earlier. In particular, the Gaussian ``kernels'', motivated by the assumed (Gaussian) distribution of the noise, can be thought of more generally as one of a myriad of choices of positive-definite kernels that can be employed to construct more general denoisers, as described below.

\subsubsection*{The General Pseudo-Linear Form }

The pseudo-linear form is very convenient for the analysis of practical denoisers in general~\cite{milanfar2013tour,takeda2006}. But even more importantly, it is a fundamental and widespread approach to denoising that decomposes the operation into two distinct steps. First is a nonlinear step where data-dependent weights $\bW(\vx,\alpha)$ are computed. Next is a \textit{linear} step where weighted averages of the input pixels yield each output pixel. More specifically, for each output pixel $x_i$, the denoiser can be described as: 
\begin{equation}
\Hat{x}_i=\sum_j W_{ij}(\vx,\alpha) \:x_j.
\end{equation}
Gathering all the weights into a matrix $\bW(\vx,\alpha)$ reveals the denoiser in {\em pseudo-linear} matrix form: 
\begin{equation} 
f(\vx,\alpha) = \mathbf{W(x,\alpha)} \:\vv{x}.
\end{equation}

Generally speaking, the weights are computed based on the affinity (or similarity) of pixels, measured using a ``kernel'' (a symmetric positive-definite function). When properly normalized, these yield the weights used to compute the output pixels as a weighted average. For instance, in the non-local means~\cite{buades_nonlocal} case
\begin{equation} \label{eq:gausskernel}
K_{ij}(\vx,\alpha) = \exp(-\|\vv{R}_{ij}\vv{x}\|^2 /2\alpha^2), \;\;\;\;\; \text{where} \;\;\;\; \|\vv{R}_{ij}\vv{x}\|^2 = \|\vx_i -\vx_j\|^2, 
\end{equation}
and $\vx_i$ denotes a patch of pixels centered at $i$. There exist many other possibilities~\cite{genton01}, a practical few of which are shown in Table~\ref{table:kernels}. 
\begin{table}
\centering
\footnotesize
\caption{Some well-known isotropic, positive-definite kernels $K(\|\bR_{ij}\vx\|,\alpha)$}
\label{table:kernels}
\begin{tabular}{l@{\hspace{2em}}l@{\hspace{2em}}l}
\hline
Name & Kernel  \\
\hline
Gaussian  & $\exp(-\|\vx_i-\vx_j\|^2/\alpha ) $   \\
Exponential & $\exp(-\|\vx_i-\vx_j\|_1/\alpha) $  \\
Cauchy & $ 1/\left(1+\alpha\|\vx_i-\vx_j\|^2\right)$  \\
\hline 
\end{tabular}
\end{table}
When normalized, these affinities give the weights $W_{ij}$ as follows 
\begin{equation}
W_{ij} = \frac{ K_{ij}}{\sum_{j}K_{ij}} \;\;\;\;\;\; \implies \;\;\;\;\; \sum_{i}W_{ij} = 1.
\end{equation}
In more compact notation\footnote{Since the weights sum to $1$ across the rows of $\bW$, this matrix is \textit{row-stochastic}. }:
\begin{equation}
\vv{W}(\vv{x},\alpha) = \mathbf{D}^{-1}(\vx,\alpha) \vv{K}(\vv{x},\alpha),
\end{equation}
where $\bD(\vx,\alpha) = \operatorname{diag}[d_1,d_2,\cdots,d_N]$ is a diagonal normalization matrix constructed from the row sums ($d_i = {\sum_{j}K_{ij}}$) of
$\bK(\vx,\alpha)$. 

\paragraph{Remark:} For common kernels such as those highlighted in the above table, the parameter $\alpha$ controls the \textit{spread} of the kernel. Therefore, as $\alpha\rightarrow 0$, the kernel approaches a scaled Dirac delta: $K_{ij}(\vx,0) = d\:\delta_{ij}$, or equivalently, the Kernel matrix is a scaled identity: $\bK(\vx,0) = d\bI$. Consequently, normalizing gives $\bW(\vx,0) = \bI$. If in addition $\bW(\vx,\alpha)$ is symmetric, then the denoiser can be approximated as the gradient of an energy (see discussion in previous Section~\ref{sec:energy-denoisers}): 
\begin{equation}
    f(\vx,\alpha) = \nabla\left[\vx^T\bW\vx\right].
\end{equation}

In practice, symmetry of the filter matrix $\bW(\vx,\alpha)$ is not a given\footnote{Though one can empirically verify that such weight matrices are approximately symmetric~\cite{romano2016little,dobigeon2024}}. Despite the fact that the kernel matrix $\bK(\vx,\alpha)$ is symmetric, the resulting weight matrix $\bW(\mathbf{x},\alpha) = \mathbf{D}^{-1}(\vx,\alpha) \vv{K}(\vv{x},\alpha)$ is not so, due to the non-trivial diagonal normalization by $\bD$. Fortunately, one can modify $\bW(\vx,\alpha)$ to satisfying the symmetry condition as detailed in
~\cite{milanfar2013symmetrizing,milanfar2016new}. This is accomplished by applying Sinkhorn balancing to $\bW$ (or equivalently to $\bK$), resulting in a symmetric and \textit{doubly}-stochastic weight matrix, which can incidentally improve mean-squared error denoising performance over the baseline - see also~\cite{wormell2021spectral}. 

Alternatively, one can take a different approach via a first-order Taylor series~\cite{milanfar2016new,localkernels2019}: 
\begin{eqnarray} \label{eq:approxSymmetry}
\bW(\vx,\alpha) \approx \bI + \beta \left(\vv{K}(\vv{x},\alpha) - \vv{D}(\vv{x},\alpha) \right),
\end{eqnarray}
where $\beta^{-1} = \frac{1}{N} \sum_{i} d_{ii}$. The right-hand side is evidently symmetric.

To give some additional context to this approach, note that when applying a filter to an image, standard practice is to normalize the filter coefficients in order to maintain the local brightness level from input to output image. This is particularly important where nonlinear filters are
concerned, where the effect on local brightness and contrast can be complex. The symmetrization approach presents a way of achieving the same level of control over the local filter behavior without the
need for this normalization.

As described in~\cite{talebi2016fast}, the approximation works better - in terms of the distortion introduced to the output image - when the diagonal entries of the matrix $\bD$ are more tightly concentrated around their mean. 
\subsection{Summary}
The takeaway message from the above discussion is that denoisers we described share some important properties in common. Namely, they have the form $f(\vx) = \vx - \alpha g(\vx)$ where $g$ is the gradient of some scalar function. Furthermore, they are:\vspace{.5em}

\noindent \textbf{Perturbation of the Identity:} The ideal behavior of a denoiser when the noise is absent ($\alpha=0$) is to give the input image back, unchanged. This is what we identified as Propertry 1 in the introductory Section~\ref{sec:intro}. We've seen that both Bayesian (MAP, and MMSE) denoisers, and their (ideal) empirical approximations satisfy this condition. \vspace{.5em}

\noindent \textbf{Shrinkage Estimators:} The general form 
$f(\vx) = \vx - \alpha v(\vx)$ can be interpreted as the ``trivial'' denoiser $\vx$ with a \emph{correction} term $\alpha v(\vx)$ that pulls the components of the noisy input toward zero. It is remarkable that these denoisers have the same form as the original James-Stein estimator~\cite{james1961}, where $\vx$ was interpreted as the maximum-likelihood estimator, and $\alpha v(\vx)$ played the role of a Bayesian 
``correction''. It has been observed~\cite{milanfar2013tour,kadkhodaie2024,srinivas2019full} that such denoisers behave (at least locally) as shrinkage operations in an orthonormal basis adapted to the underlying structure of the image.\vspace{.5em}

\noindent \textbf{Gradient Descent on Energy:} We noted that many denoisers can be written in the form $f(\vx) = \vx - \alpha \nabla \mathcal{E}(\vx)$. It is obvious that the right-hand side defines one step in a steepest descent iteration. Repeated applications of a denoiser have the effect of marching toward a local stationary point of the energy.\vspace{.5em}

\noindent \textbf{Approximate Projection:} It has been pointed out elsewhere~\cite{permenter2024} that if we accept the assertion that real-world images with $N$ pixels are approximately contained in low-dimensional manifolds of $\mathbb{R}^N$~\cite{carlsson2009}, then adding noise is equivalent to orthogonal perturbation away from the manifold, and denoising is approximately a projection onto the manifold. In particular, for small noise, denoising is precisely a projection onto the local tangent of said manifold. As such, the work of denoising is essentially analogous to manifold learning.

\section{Denoising, the Score Function, and Generative Modeling} 
A crucial link between denoising and the score function enables denoisers to learn complex probability distributions. In modeling real-world data, and images in particular, we are typically faced with a complex, high-dimensional probability density, $P(\cdot)$. Explicitly modeling such a distribution can be computationally intractable or extremely difficult. The score function, defined as the gradient of the log probability density, can provide a way through. 
\begin{equation}
    s(\vx,\alpha) = \nabla \log P(\vx,\alpha).
\end{equation} 

Instead of modeling the distribution $P(\vx,\alpha)$ directly, we can learn, or approximate, the score function~\cite{song2019generative}. Denoising techniques are a way to implicitly learn the score function roughly as follows: an estimate of the score function around a ``clean'' image is obtained by corrupting it with noise, training a model to reconstruct the original clean image from the noisy version, and measuring the denoising \textit{residual}:
\begin{equation} \label{eq:score_approx}
    -s(\vx,\alpha) \approx \frac{\vx - f(\vx,\alpha)}{\alpha}.
\end{equation} 

At first blush, it is not at all clear why this is a reasonable procedure. Yet there are a number of ways~\cite{vincent2011connection,sohl2015deep,song2019generative,ho2020denoising,song2021scorebased} to motivate this idea -perhaps none more direct than by using \textit{Tweedie's formula} introduced earlier in Eq. \eqref{eq:tweedie}: 
\begin{equation}
\vv{\Hat{x}}_\mathit{mmse} = \vx + \alpha \nabla \log P(\vx,\alpha).
\end{equation}
Rewriting this establishes a direct and \textit{exact} relationship between score function and the MMSE denoiser:
\begin{equation}\label{eq:score}
-s(\vx,\alpha) = \frac{\vx - \vv{\Hat{x}}_\mathit{mmse}}{\alpha}.
\end{equation}
Despite its elegance, the MMSE estimator is typically difficult to compute, or entirely inaccessible. Therefore as a proxy, often other denoisers are used, which may only be rough approximations of the MMSE (Eq. \eqref{eq:score_approx}).

One can take a broader point of view by considering \textit{ideal} denoisers:
\begin{equation}
f(\vx,\alpha) = \nabla \mathcal{E}_0(\vx,\alpha),
\end{equation}
where $\mathcal{E}_0(\vx,\alpha)$ is of the form
\begin{equation}\label{eq:energy}
\mathcal{E}_0(\vx,\alpha) = \frac{1}{2}\|\vx\|^2 - \alpha\mathcal{E}(\vx,\alpha).
\end{equation}
Energy functions such as these can be learned~\cite{song2019generative,salimans2021,song2021,cohen2021}, and the resulting denoisers have the appealing form\footnote{We remind the reader this includes the MMSE and MAP, but can be more general.}: 
\begin{equation}
f(\vx,\alpha) = \vx - \alpha \nabla \mathcal{E}(\vx,\alpha).
\end{equation}
Or equivalently
\begin{equation}
\nabla \mathcal{E}(\vx,\alpha) = \frac{\vx - f(\vx,\alpha)}{\alpha}.
\end{equation}
This illustrates again that the energy function is a proxy~\cite{salimans2021} for the score $s(\vx,\alpha) \approx -\nabla \mathcal{E}(\vx,\alpha)$, and the resulting denoiser's residual can be used as an approximation of the score.

\subsection{Denoising as the Engine of Diffusion Models}

\begin{figure}[th]
    \centering
    \includegraphics[width=0.8\linewidth]{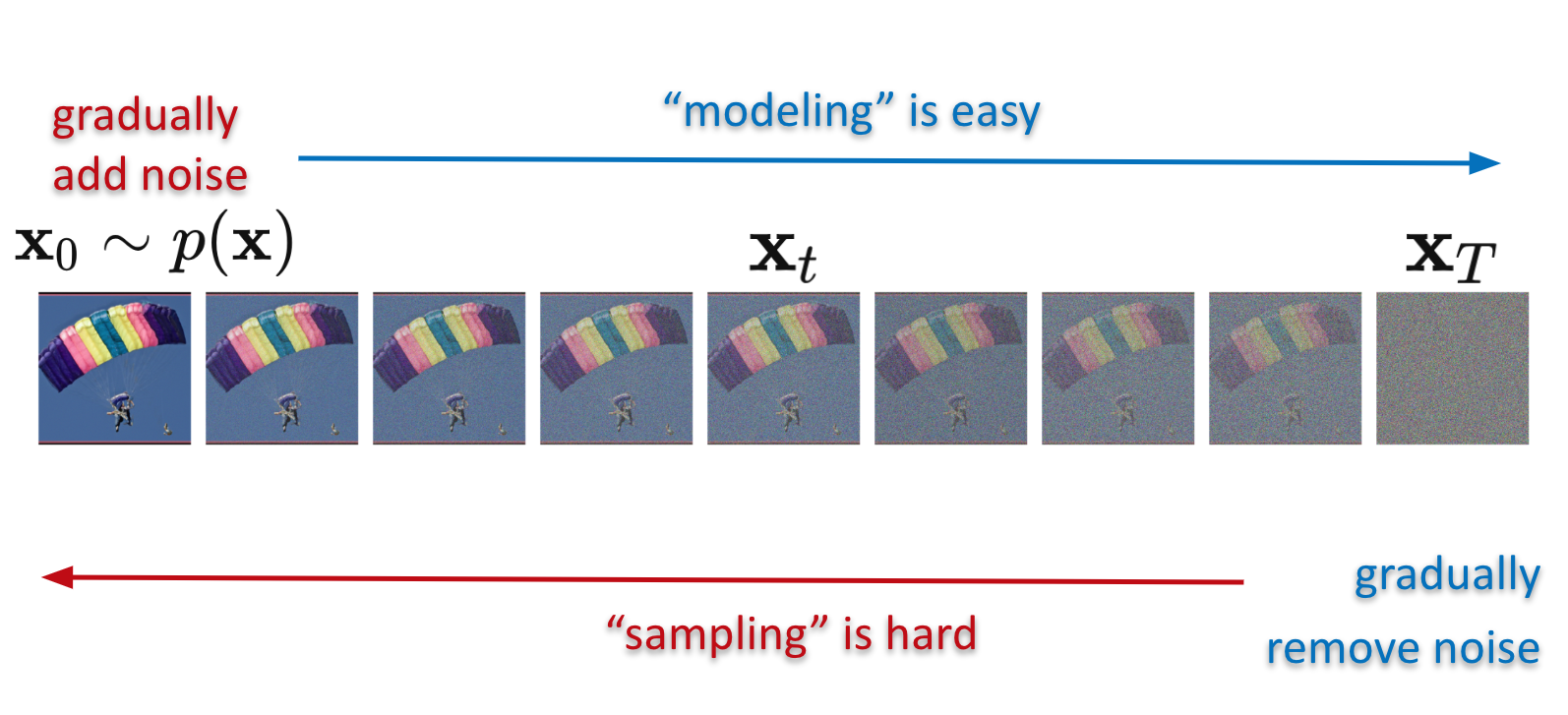}
    \caption{Diffusion: Forward (Modeling) and Backward (Sampling).}
    \label{fig:Diffusion}
\end{figure}

Denoising Diffusion and Flow generative models~\cite{sohl2015deep,song2019generative,Ho2020,song2021denoisingImplicit,song2021scorebased,lipman2023flow} have become an important area of research in generative modeling.
They operate by progressively corrupting training data with noise until it's indistinguishable from random noise, then learning to systematically reverse this corruption. By training a model to iteratively denoise, it gains the ability to generate entirely new, coherent data samples from a starting point of pure noise, effectively converting noise into meaningful structures like images or other data forms (Figure~\ref{fig:Diffusion}). 

Despite their popularity, expressive power, and tremendous success in practice, there's been relatively little intuitive clarity about how they operate. At their core, these models enable us to start with a sample from one distribution (e.g. a Gaussian), and arrive (at least approximately) at a sample from a target distribution $P(\vx,\alpha)$. But how is this magic possible? Referring to Figure~\ref{fig:Diffusion}, let's say we begin with a sample $ \vx_T \sim \mathcal{N}(0,\alpha \mathbf{I})$, where $\alpha \gg \text{Var}[\vx]$. 

One simple way to activate this sampling process is to directly consider a \textit{flow} differential equation
\begin{equation} 
    \frac{d\vx_t}{dt} = -\frac{1}{2}\frac{d \alpha_t}{dt} \nabla \log P(\vx_t, \alpha_t),
    \label{eq:probflow}
\end{equation}
where the right-hand side is the score function introduced earlier, and $\alpha_t$ is the noise level at time $t$. This differential equation, called a \textit{probability flow}~\cite{song2021scorebased}, by construction moves the initial condition gradually toward the distribution $P(\vx,\alpha)$. Solving this equation requires (a) selecting a numerical scheme, and (b) having access to the score function. 

If we have access to an MMSE denoiser at every $t$, we can invoke Tweedie's formula to write:
\begin{align}
    \frac{d\vx_t}{dt} &= \frac{1}{2}\frac{d\alpha_t }{dt} \frac{\left(\vx_t-\mathbb{E}[\vx_0|\vx_t]\right)}{\alpha_t},
    \label{eq:residual_flow}
\end{align}
\noindent which we call a \textit{residual} flow. As we've described in the previous sections, lack of access to the MMSE denoiser forces us to select a different denoiser and therefore solve only an approximate version of the desired Eq.~\eqref{eq:probflow}.\\

\noindent \textbf{Understanding the Velocity Coefficient.} A key question arises: How is the velocity coefficient (the term multiplying the \emph{residual}) in \eqref{eq:residual_flow} ODE determined? Let's assume the process has a conditional variance $\text{Var}[\vx_t | \vx_0] = \alpha_t$ at time $t$ (i.e., noise level). The ODE is then constructed such that this variance evolves consistently, meaning $\text{Var}[\vx_{t - dt} | \vx_0] = \alpha_{t-dt}$.

A first order discretization of \eqref{eq:residual_flow} yields:
\begin{align}
\vx_t - \vx_{t-dt} &= \frac{\alpha_t - \alpha_{t-dt}}{2 \alpha_t}\left(\vx_t-\mathbb{E}[\vx_0|\vx_t]\right) \\
\vx_{t-dt} &= \frac{\alpha_t + \alpha_{t-dt}}{2 \alpha_t}\vx_t - \frac{\alpha_t - \alpha_{t-dt}}{2 \alpha_t}\mathbb{E}[\vx_0|\vx_t]. 
\end{align}

This allows us to derive the conditional variance of $\vx_{t-dt}$ given $\vx_0$:
\begin{align}
\text{Var}[\vx_{t-dt}|\vx_0] &= \text{Var}\left[\frac{\alpha_t + \alpha_{t-dt}}{2 \alpha_t}\vx_t - \frac{\alpha_t - \alpha_{t-dt}}{2 \alpha_t}\mathbb{E}[\vx_0|\vx_t] \bigg| \vx_0 \right] \\
&= \frac{(\alpha_t + \alpha_{t-dt})^2}{4\alpha_t^2}\text{Var}[\vx_{t}|\vx_0]  \\
&= \frac{(\alpha_t + \alpha_{t-dt})^2}{4\alpha_t^2} \cdot \alpha_t \\
&= \alpha_{t-dt} + \frac{(\alpha_{t-dt} - \alpha_t)^2}{4\alpha_t} \\
&\approx \alpha_{t-dt},
\end{align}
where the final approximation holds for small $dt$. This demonstrates that the velocity coefficient in \eqref{eq:residual_flow} effectively ensures the consistent evolution of the conditional variance, crucial for accurately capturing the underlying process dynamics.

A crucial point for our discussion is that the probability flow described by equation \eqref{eq:probflow} can be proven to yield the same marginal distributions as the stochastic formulation presented in~\cite{song2021scorebased}. This implies that, in the limit, if we initialize with samples from a Gaussian distribution, the solution is guaranteed to produce samples that match the data distribution.
While a comprehensive mathematical analysis of diffusion models is beyond the scope of this work, we encourage interested readers to delve into the foundational works~\cite{sohl2015deep,song2019generative,Ho2020,song2021denoisingImplicit,song2021scorebased} or the excellent introductory overviews~\cite{weng2021diffusion,dieleman2023perspectives,chan2024tutorial} for a deeper understanding.

\section{Denoisers in the Context of Inverse Problems}

Consider the following formulation of a linear inverse problem\footnote{While we'll only described linear inverse problems in this exposition, the RED and other frameworks are equally applicable to nonlinear inverse problems (e.g.,~\cite{Wu2019}), albeit with the caveat that some of the nice convexity properties of the overall loss no longer hold.}: The data is given by the following model 
\begin{equation}
\vy = H\vx + \be,
\label{eq:linear-inverse-problem}
\end{equation}
where $H \in \mathbb{R}^{m \times n}$ is the forward operator (e.g., degradation or measurements operator), $\be \in \mathbb{R}^m$ is additive white Gaussian noise, and the task is retrieving $\bx \in \mathbb{R}^n$ from $\by \in \mathbb{R}^m$.

A nominal solution can be obtained by solving this optimization problem:
\begin{equation} \label{eq:inverse-problem}
    \hat{\vx} = \argmin{\vx} l(\vy,\vx) + \lambda \bR(\vx,\alpha),
\end{equation}
\noindent where $l(\vy,\vx) = \frac{1}{2}\|H\vx -\vy\|^2$ captures the Gaussian nature of the noise, and $\bR(\cdot)$ is a regularization term intended to stabilize the solution, and $\lambda>0$ is a regularization parameter. 

Over the last several decades, a vast number of choices for the regularizer $\bR(\vx,\alpha)$ have been proposed with varying degrees of success. Early approaches often relied on hand-designed priors to encourage desired properties in the solution, such as sparsity or smoothness~\cite{rudin1992nonlinear,daubechies2004iterative,candes2006robust,donoho2006compressed}. 
Iterative Shrinkage/Thresholding (IST) algorithms~\cite{figueiredo2003algorithm, daubechies2004iterative, figueiredo2005bound, combettes2005signal, hale2007fixed} utilize the shrinkage/thresholding function (Moreau proximal mapping) derived from the regularizer $\bR$~\cite{combettes2005signal} to solve optimization problems. However, the non-smoothness of many regularizers and the scale of these problems pose computational challenges. Proximal methods like FISTA~\cite{beck2009fast} and ADMM~\cite{eckstein1992douglas, afonso2010fast} present more efficient solutions by leveraging the proximal operator, which can be interpreted as applying a denoising step to intermediate solutions.

More recently, and independently of the machine learning literature, a fascinating connection has emerged between denoising algorithms and inverse problems. Powerful denoising algorithms, particularly those leveraging deep learning, have been shown to implicitly encode strong priors about natural signals. By incorporating these denoisers into the optimization framework, we can effectively leverage their learned priors to achieve state-of-the-art performance in various inverse problems
~\cite{plugandplay, chan2016plug, brifman2016turning, teodoro2016image,romano2016little,cohen2021regularization,arridge2019solving,laumont2022bayesian,hurault2022proximal,kamilov2023plug}. This approach effectively blurs the lines between traditional regularization techniques and modern denoising methods, offering a new paradigm for solving inverse problems.

Learning priors from data has a long history starting in the statistical literature with the concept of ``empirical Bayes''(see e.g.~\cite{robbins56,efron2012large}). More recently, both implicit and explicit methods have been developed to learn the distribution of images~\cite{elad2010sparse,zoran2011learning,yu2011solving,aguerrebere2017bayesian,holden2022bayesian,altekruger2023patchnr}. In particular, the vast recent literature on diffusion models is all about mapping a known distribution (typically a multidimensional Gaussian) to an empirical distribution (learned from collections of images in a desired domain)
~\cite{song2019generative,Ho2020,song2021scorebased,lipman2023flow}.

As we described earlier, access to a high quality denoising engine affords us the possibility to learn, or at least locally approximate, the geometry of the image manifold. This approximate geometry is learned based on a residual: the difference between a \textit{noisy} image and its denoised version. This enables us to formulate inverse problems as general optimization tasks, where the denoiser (or more specifically a functional based on it) is used as a regularizer. 

In order to solve the optimization problem~\eqref{eq:inverse-problem}, it is necessary to evaluate the gradient of the objective, which is as follows: 
\begin{equation} \label{eq:inverse-problem-grad}
     H^T(\vy-H\vx) + \lambda \nabla \bR(\vx,\alpha).
\end{equation}
A key concern is how to compute $\nabla \bR(\vx,\alpha)$. In this respect, classical choices of $\bR(.)$ such as $L_p$ norms have been fairly convenient and successful; but also shown to have limited power to model natural images~\cite{elad2010sparse}. 

Another choice that has proved more effective is (image-adaptive) \textit{Laplacian} regularizers~\cite{elmoataz2008,coifman_geometric,kheradmand2014general,romano2015boosting} that implicitly contain a (pseudo-linear) denoiser inside. Namely,   
\begin{equation}
    \bR(\vx,\alpha) = \frac{1}{2} \vx^T\; \bL(\vx,\alpha)\; \vx.
\end{equation}
In~\cite{romano2016little}, we developed a natural extension of this idea called \textit{Regularization by Denoising} (RED), where the regularizer is constructed from a more general denoiser $f(\vx,\alpha)$:
\begin{equation}
    \bR_{red}(\vx,\alpha) = \frac{1}{2} \vx^T(\vx- f(\vx,\alpha)).  
\end{equation}
Note the intuition behind this prior: the value of $\bR_{red}(\vx,\alpha)$ is low if the cross-correlation between the image and the denoising residual is small, or if the residual itself is small due to $\vx$ being a fixed point of $f(\cdot)$.

But with this generality comes a challenge: can the gradient of the regularizer be computed easily? The answer is yes, when $f(\vx,\alpha)$ is ideal and locally homogeneous~\cite{romano2016little}. This is not difficult to prove:
\begin{eqnarray}
       \nabla \vx^T(\vx- f(\vx,\alpha)) & = & 2\vx - \nabla\left[\vx^Tf(\vx,\alpha)\right] \\
     & = & 2\vx - f(\vx,\alpha) - \nabla f(\vx,\alpha)\vx \\
     & = & 2\vx - 2f(\vx,\alpha),
\end{eqnarray}
\noindent where the second line follows from the Jacobian symmetry of ideal denoisers; and the third line follows from local homogeneity and the definition of directional derivative~\cite{romano2016little}: 
\begin{align} \label{eq:homogeneous}
    \nabla f(\vx)\vx = & \lim_{\epsilon\rightarrow 0} \frac{f(\vx+\epsilon \vx) - f(\vx)}{\epsilon}  \\
      = & \lim_{\epsilon\rightarrow 0} \frac{(1+\epsilon)f(\vx) - f(\vx)}{\epsilon} = f(\vx).
\end{align}
Replacing $\nabla \bR_{red}(\vx,\alpha) = \vx- f(\vx,\alpha),$ 
for the gradient in \eqref{eq:inverse-problem-grad}, we have the following expression for the gradient of the objective: 
\begin{equation}
     H^T(\vy - H\vx) + \lambda(\vx- f(\vx,\alpha)).
\end{equation}
The most direct numerical procedure for solving this equation is a fixed point iteration that sets the gradient of the objective to zero: 
\begin{equation} 
 H^T(\vy - H\vx_{k+1} ) +  \lambda (\vx_k- f(\vx_k,\alpha)) = 0.
\end{equation}
Equivalently,
\label{eq:fixed-point}
\begin{align}
    \vx_{k+1} & = \bb + \bM\;f(\vx_k,\alpha),
\end{align}    
where
\begin{align}
    \bb = \left[H^TH +\lambda I\right]^{-1}H^T \vy, \quad \,
    \bM = \lambda \left[H^TH + \lambda I\right]^{-1}.
\end{align}
Here, $\bb$ is the (fixed) linear \textit{pseudo-inverse} solution and $\bM$ is also a fixed matrix. Procedurally, we start with $\vx_{0} = \vy$, denoise it, and then a linear operator $\bM$ is applied and a bias $\bb$ is added - this leads to an updated estimate, and the process is repeated. Note that the structure of this iterative process is not altogether different from a denoising diffusion process
~\cite{Ho2020} where a denoiser is repeatedly applied. In fact, when $H = I$ we see the structure of a \textit{bridge} diffusion process~\cite{delbracio2023inversion}: 
\begin{align}
    \vx_{k+1} & = \frac{1}{1+\lambda} \vy+ \frac{\lambda}{1+\lambda} f(\vx_k,\alpha).
\end{align}

In a general statistical setting, the scalar valued $\bR(\vx,\alpha)$ is often the result of assuming a prior whose negative-log is \textit{interpreted} as the regularizer: 
\begin{equation}
  \bR(\vx,\alpha) = -\log P(\vx,\alpha).
\end{equation}
However, in cases where a denoiser is used to \textit{construct} a regularizer, the role of the regularizer $\bR(\vx,\alpha) \geq 0$ is that of an energy function that we implicitly use to define a Gibbs distribution~\cite{blake2011}:
\begin{equation}
 P(\vx,\alpha) \propto \exp{\left[-\bR(\vx,\alpha)\right]}.
\end{equation}
In the particular case of RED: $\bR_{red}(\vx,\alpha) = \frac{1}{2} \vx^T(\vx- f(\vx,\alpha))$, Equation~\eqref{eq:homogeneous} implies that an ideal and locally homogeneous denoiser has the form $f(\vx) = \nabla f(\vx) \vx$, which means that under these conditions, the RED regularization can be thought of as a (\textit{pseudo-quadratic}) energy function: 
\begin{equation}
    \bR_{red}(\vx,\alpha) = \frac{1}{2} \vx^T\nabla f(\vx,\alpha) \vx.
\end{equation}

\subsection*{Posterior Sampling with Denoisers} 
An alternative approach to solving inverse problems is to leverage pretrained denoisers as priors for generating samples from the posterior distribution~\cite{kadkhodaie2021stochastic,kawar2022denoising,chung2023diffusion}. Given measurements $\vy$, our goal is to generate samples $\vx$ that follow the distribution $P(\vx|\vy)$, where the prior $P(\vx)$ is implicitly defined by the denoiser.

To achieve this, we can adapt the generative sampling strategy from Equation \eqref{eq:probflow} to sample from the posterior distribution $P(\vx|\vy)$ instead of the prior $P(\vx)$:
\begin{align} 
    \frac{d\vx_t}{dt} &= -\frac{1}{2}\frac{d \alpha_t}{dt} \nabla \log P(\vx_t|\vy, \alpha_t) \\
    &=  -\frac{1}{2}\frac{d \alpha_t}{dt} \left(\nabla \log P(\vx_t, \alpha_t) + \nabla \log P(\vy|\vx_t, \alpha_t)\right),
    \label{eq:probflow_posterior}
\end{align}
starting from $\vx_T\sim P(\vx_T)$. The second equality is given by Bayes rule. 

We recognize the first term as the score function $\nabla \log P(\vx_t, \alpha_t)$, which can be connected to the MMSE denoiser through Tweedie's formula~\eqref{eq:tweedie}. The second term in~\eqref{eq:probflow_posterior} quantifies how well the current sample $\vx_t$ explains the measurements $\vy$, but this is generally intractable to compute.

\vspace{.5em}
\noindent \textbf{Diffusion Posterior Sampling framework (DPS):}
One approach to address this intractability is the Diffusion Posterior Sampling framework~\cite{chung2023diffusion}.  DPS approximates the intractable term with $\log P(\vy|\vx_t,\alpha_t) \simeq \log P(\vy|\mathbb{E}[\vx_0|\vx_t],\alpha_t)$, based on the assumption that $p(\vx_0|\vx_t) \simeq \delta(\vx_0 - \mathbb{E}[\vx_0|\vx_t])$.

Considering a linear measurement model as in Equation~\eqref{eq:linear-inverse-problem}, this approximation leads to:
\begin{align}
\log P(\vy|\vx_t)  \simeq -\| H\mathbb{E}[\vx_0|\vx_t] - \vy\|^2.
\end{align}
Substituting this into~\eqref{eq:probflow_posterior}, we obtain:
\begin{align}
\frac{d\vx_t}{dt} =  \frac{1}{2}\frac{d \alpha_t}{dt} \left(\frac{\vx_t-\mathbb{E}[\vx_0|\vx_t]}{\alpha_t} + \rho_t \nabla_{\vx_t} \| H\mathbb{E}[\vx_0|\vx_t] - \vy||^2\right),
\end{align}
where $\rho_t$ is a hyperparameter balancing the influence of the prior and the measurements.  In practice, we utilize a denoiser network $f(\vx_t, \alpha_t)$ to approximate the conditional expectation $\mathbb{E}[\vx_0|\vx_t]$.

\vspace{.5em}
\noindent \textbf{Growing Importance of Denoising Diffusion Models:} Denoising diffusion models are rapidly emerging as a powerful tool for solving inverse problems across various domains.  This success often stems from combining the strengths of diffusion models with additional approximations or specialized techniques.  A growing body of research explores these approaches~(\cite{jalal2021robust,song2021scorebased,kawar2021snips,kadkhodaie2021stochastic,kawar2022denoising,laumont2022bayesian,zhu2023denoising,song2023pseudoinverseguided,feng2023score,mardani2024variational,chung2024prompt,rout2024solving,wu2024principled}; see~\cite{daras2024survey} for a comprehensive review).

\section{Conclusions}
In this paper, we have explored the multifaceted nature of denoising, showcasing its far-reaching impact beyond the traditional task of noise removal. We have highlighted the structural properties of denoisers, their connection to Bayesian estimation and energy-based models, and their ability to act as powerful priors and regularizers in various applications. The surprising effectiveness of denoisers in tasks from generative modeling to inverse problems underscores their versatility and potential for future research. The continued evolution of denoising techniques, coupled with advancements in machine learning, promises to unlock even more innovative applications and deeper insights into the underlying structure of images.

\subsubsection*{Acknowledgments}
The authors extend their sincere gratitude to Mojtaba Ardakani, Michael Elad, Vladimir Fanaskov, Mario A. T. Figueiredo, Ulugbek Kamilov, Jos\'e Lezama, Ian Manchester and Miki Rubinstein  for their valuable feedback.

We also extend our sincere thanks to the vast research community whose dedication over decades has driven remarkable progress in denoising. The advancements in this field are a testament to collective effort, and it is beyond the scope of this work to fully acknowledge the extensive and diverse body of literature on this topic.

\bibliographystyle{IEEEtran}
\bibliography{paper}

\end{document}